%% file: DeepLearningKitPaper.tex
\pdfoutput=1
\documentclass[xcolor=table]{article} 
\usepackage{nips13submit_e,times}
\usepackage{hyperref}
\usepackage{url}
\usepackage[final]{graphicx}

\usepackage{listings}
\usepackage{xcolor}
\usepackage{amsmath}
\usepackage{algorithm}
\usepackage[noend]{algpseudocode}

\makeatletter
\def\BState{\State\hskip-\ALG@thistlm}
\makeatother

\title{DeepLearningKit - an GPU Optimized Deep Learning Framework for Apple's iOS, OS X and tvOS developed in Metal and Swift}
\author{
Amund~Tveit\thanks{\url{http://DeepLearningKit.org}} \\
\href{http://deeplearningkit.org}{DeepLearningKit}\\
\texttt{opensource@deeplearningkit.org} \\
\And
Torbj{\o}rn Morland* \\
\href{http://deeplearningkit.org}{DeepLearningKit}\\
\And
Thomas Brox R{\o}st*\\
\href{http://atbrox.com}{Atbrox}\\
}

\nipsfinalcopy 

\begin{document}

\maketitle

\begin{abstract}
  In this paper we present DeepLearningKit - an open source framework
  that supports using pre- trained deep learning models (convolutional
  neural networks) for iOS, OS X and tvOS. DeepLearningKit is
  developed in Metal in order to utilize the GPU efficiently and Swift
  for integration with applications, e.g. iOS-based mobile apps on
  iPhone/iPad, tvOS-based apps for the big screen, or OS X desktop
  applications. The goal is to support using deep learning models
  trained with popular frameworks such as Caffe, Torch, TensorFlow,
  Theano, Pylearn, Deeplearning4J and Mocha. Given the massive GPU
  resources and time required to train Deep Learning models we suggest
  an App Store like model to distribute and download pretrained and
  reusable Deep Learning models.
\end{abstract}

\input{introduction}

\input{modelappstore}

\input{modelimporter}

\section{Conclusion}
Have done a presentation of DeepLearningKit GPU accelerated Deep
Learning for Metal/Swift and presented directions for how it can be ported/adapted to
OpenCL/Vulkan SPIR-V.

\newpage

\small{
\bibliography{DeepLearningKitPaper}
}

\end{document}

%% file: introduction.tex
\section{GPU Accelerated Deep Learning Library}
The Metal programming language is most the efficient way of utilizing
the GPU on Apple's iOS since 2014 \cite{2014AppleMetalOverviewSandmel,2014AppleMetalFundamentalsSchreyer,2014AppleMetalAdvancedGokhan, 2014MemkiteMetalGPGPUAccelerateTveit} and OSX since 2015 \cite{2015AppleWhatsNewInMetal1,2015AppleWhatsNewInMetal2,2015MemkiteMetalOSXTveit}.
This paper gives a brief overview of a Metal and Swift based Deep Learning library named {\bf \href{http://deeplearningkit.org/}{DeepLearningKit}}, in particular
parts of Metal convolutional neural network operators for the
GPU. DeepLearningKit supports on-device Deep Learning on Apple's iOS, OS X and tvOS.

\begin{figure}[h!]
\begin{center}
\includegraphics[width=12cm]{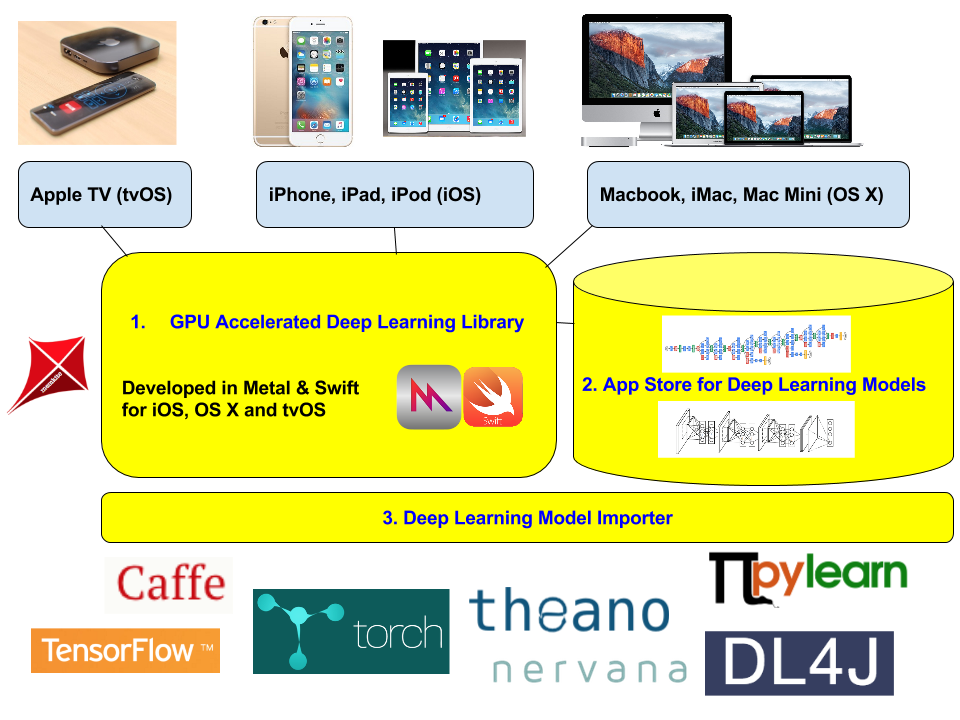}
\caption{DeepLearningKit Overview}
\label{memkiteiosstack}
\end{center}
\end{figure}

DeepLearningKit currently has shader functions for convolutional neural
networks implemented in Metal and parallelized for the GPU, operators
include: convolution, pooling, rectifier layer and softmax. In terms of
deep learning model supported it has support for Min Lin's Caffe-trained
Network In Network\cite{DBLP:journals/corr/LinCY13} (NIN - trained on CIFAR-10,
CIFAR-100 and ImageNet data sets). We also have preliminary support running
Theano\cite{bergstra+al:2010-scipy} trained LeNet (trained on MNIST
digit classification dataset). The reason we have chosen NIN is
that the network is small compared to other deep convolutional neural
networks, but at the same time provide very high classification
accuracy on images, e.g. better than AlexNet. GoogleLeNet (winner of
Imagenet 2014) uses a similar approach as NIN\cite{DBLP:journals/corr/SzegedyLJSRAEVR14}.  NIN can perhaps also
be used in non-image domains, e.g speech
recognition\cite{2015DLSpeechRecoTveit} or natural language
processing\cite{2015DLNLPTveit}. In particular one could attempt to adapt Zhang and Lecun's
encoding and 1D convolutional operators in ``Text Understanding from
Scratch''\cite{DBLP:journals/corr/ZhangL15} and use it with NIN.

\subsection{Experiences with PowerVR G6430/GT7600 on iPhone 5S/6S}
The performance of DeepLearningKit Deep Learning going from iPhone 5S (with
PowerVR G6430 according to \href{http://www.anandtech.com/show/7335/the-iphone-5s-review/7}{The iPhone 5S Review (AnandTech)}) to iPhone 6S (with PowerVR GT7600 according to \href{http://www.gsmarena.com/iphone_6s_plus_vs_s6_edge_plus-review-1324p7.php}{Apple iPhone 6S Plus vs. Samsung Galaxy S6 Edge+})
– we got 1 order of magnitude in improved performance. Calculation time
to run through a 20 layer deep convolutional neural network model for
image recognition went from approximately 2 seconds to less than 100
milliseconds. The network we we used was NIN network trained on
CIFAR-10. Based on XCode profiling we suspect that the Metal compute
drivers for the GPU weren't fine tuned, so with lower level tools (e.g. for OpenCL/Vulkan SPIR-V) for
tuning for the GPU we could probably improve performance quite a bit.

(Note that 100 milliseconds or in other words 0.1 seconds is what
Jacob Nielsen stated is one of 3 important response times – that a
user feels a system reacts instantenously)

\subsection{Effort needed to port from Metal/Swift to OpenCL/Vulkan Compute SPIR-V}
\begin{figure}[h!]
\begin{center}
\includegraphics[width=12cm]{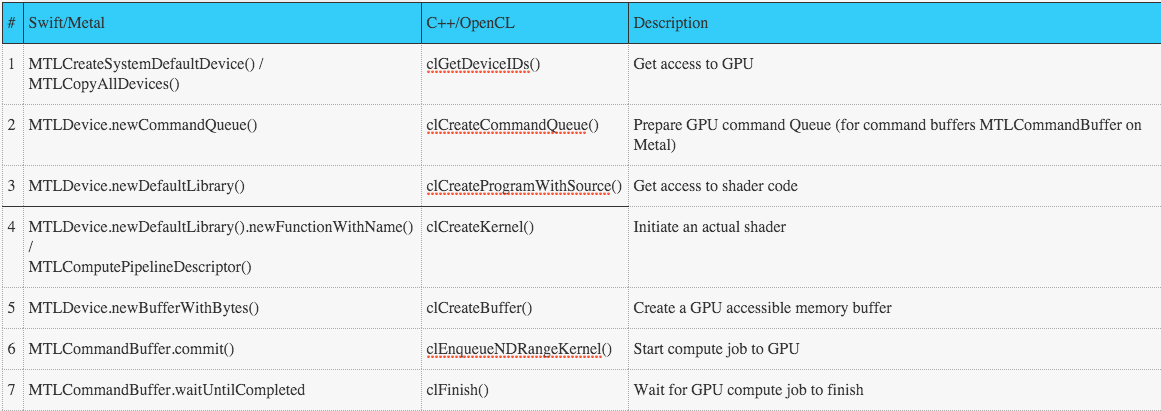}
\caption{DeepLearningKit Overview}
\label{metalopencl}
\end{center}
\end{figure}

Code needed to set up and run deep learning on the GPU, load/save
data, and setup the deep learning pipeline (convolutional neural
network)is done is done in Swift (for easy app integration on iOS, OS
X and tvOS), but can be moved to a language of selection (e.g. Java on
Android or C++/C on other devices). The Swift API for setting up Metal
resembles the corresponding OpenCL C API as shown in Figure \ref{metalopencl}.

\begin{figure}[h!]
\begin{center}
\includegraphics[width=12cm]{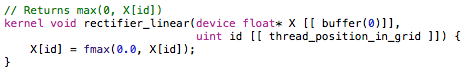}
\caption{Rectifier Function in Metal}
\label{metalcode}
\end{center}
\end{figure}

\begin{figure}[h!]
\begin{center}
\includegraphics[width=12cm]{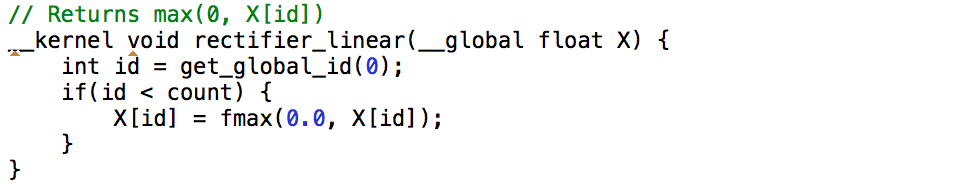}
\caption{Rectifier Function in OpenCL}
\label{openclcode}
\end{center}
\end{figure}

\begin{figure}[h!]
\begin{center}
\includegraphics[width=12cm]{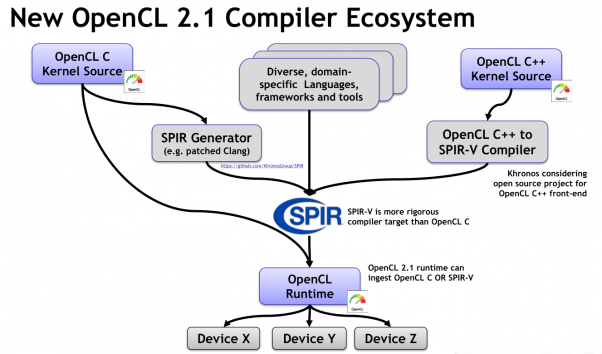}
\caption{OpenCL - Vulkan SPIR-V relationship}
\label{vulkanspir}
\end{center}
\end{figure}

The Deep Learning GPU code (e.g. shader functions with calculations of
convolution etc) is written in Metal, a language that is a subset
C++-11 and also has its own (relatively few) additions compared to
C++11. Porting the Metal code GPU code to OpenCL should be relatively
straight forward since OpenCL is also a subset of C++, as an example see figures \ref{metalcode} and \ref{openclcode} for
a rectifier function written in both Metal and OpenCL. Going from OpenCL to Vulkan SPIR-V can be done with compiler (figure \ref{vulkanspir}) for further profiling and optimization.

\begin{figure}[h!]
\begin{center}
\includegraphics[width=12cm]{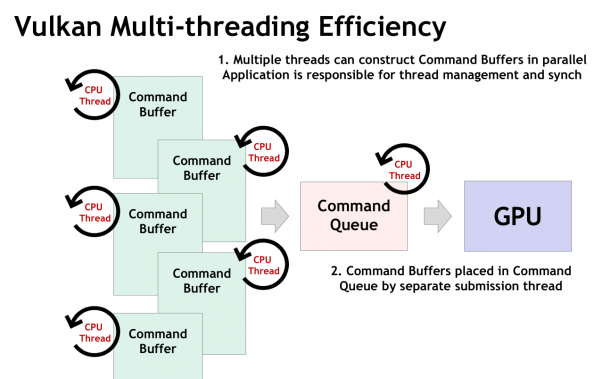}
\caption{Vulkan Multi Threading}
\label{vulkanthread}
\end{center}
\end{figure}

The threading model supported by Vulkan is 1-1 with what is developed in DeepLearningKit with Metal (figure \ref{vulkanthread}), so that should not be an issue (The equivalent classes to what Vulkan has in the figure in Metal is from left to right MTLCommandBuffer, MTLCommandQueue and MTLDevice)

\subsection{Roadmap for Deep Learning for OpenCL/Vulkan (or Metal)}
Here follows a brief overview of things we are working on or is on our roadmap.

\begin{figure}[h!]
\begin{center}
\includegraphics[width=12cm]{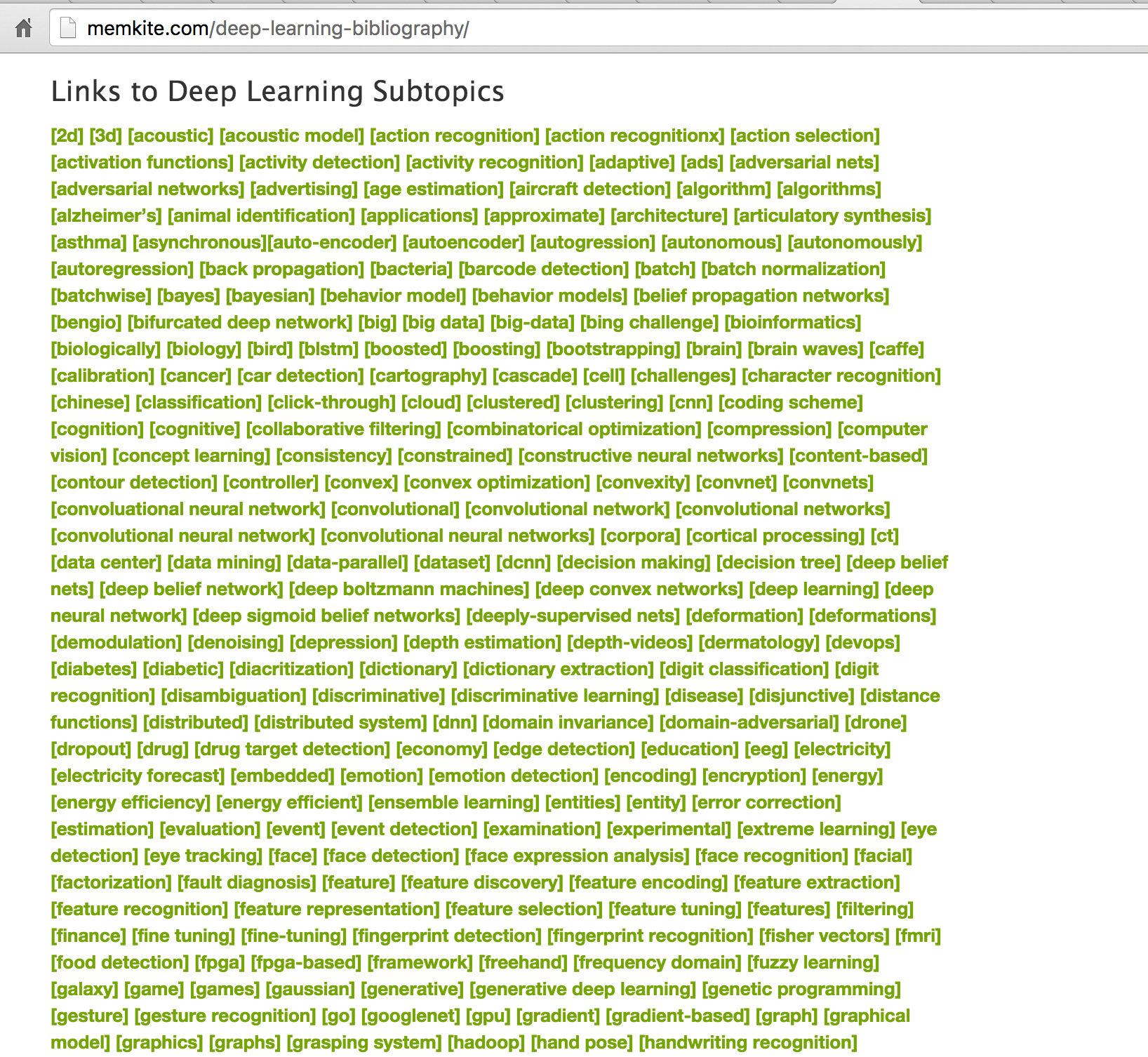}
\caption{Deeplearning.University - Keywords part 1}
\label{dluniv1}
\end{center}
\end{figure}

\begin{figure}[h!]
\begin{center}
\includegraphics[width=12cm]{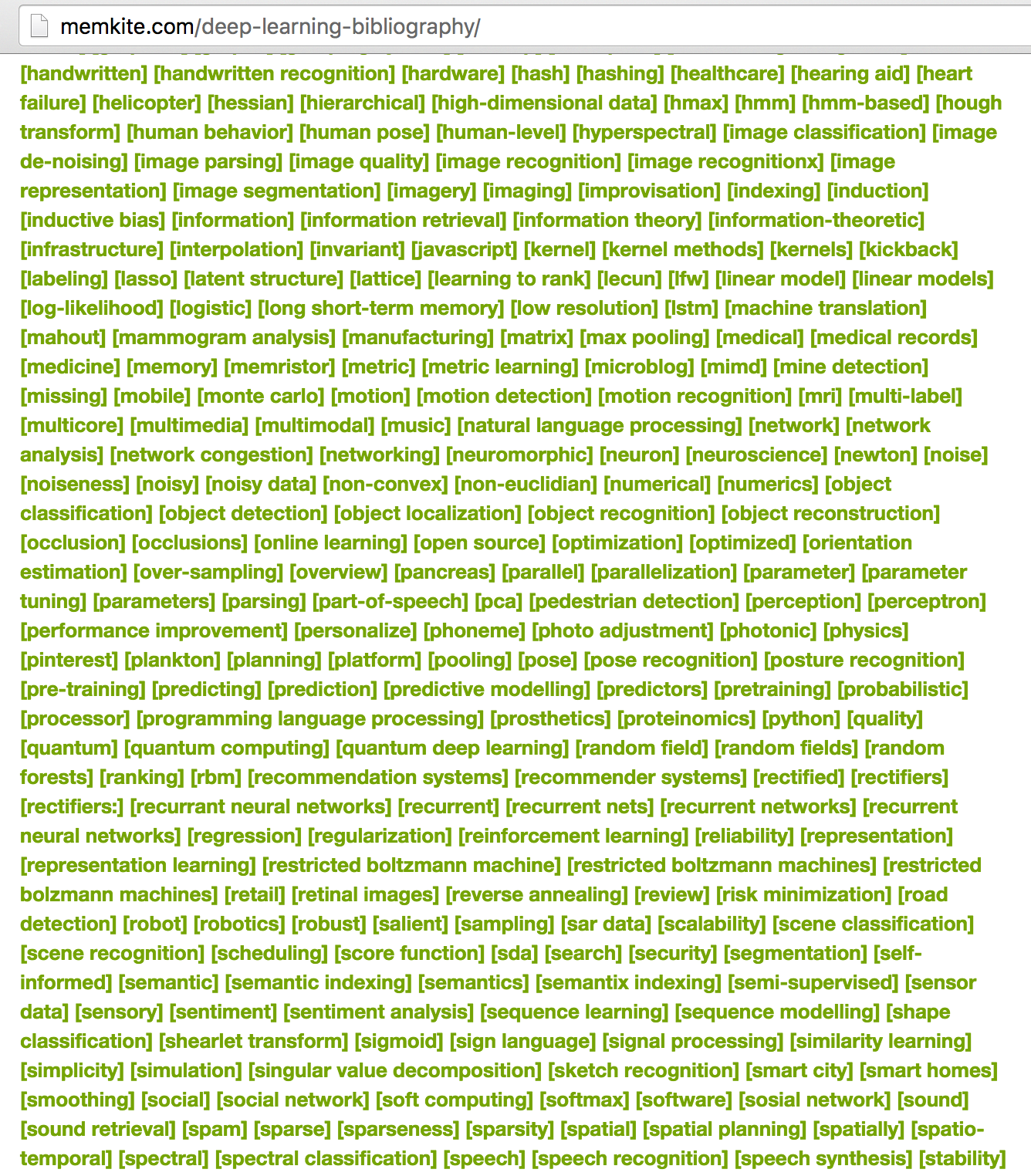}
\caption{Deeplearning.University - Keywords part 2}
\label{dluniv2}
\end{center}
\end{figure}

\begin{figure}[h!]
\begin{center}
\includegraphics[width=12cm]{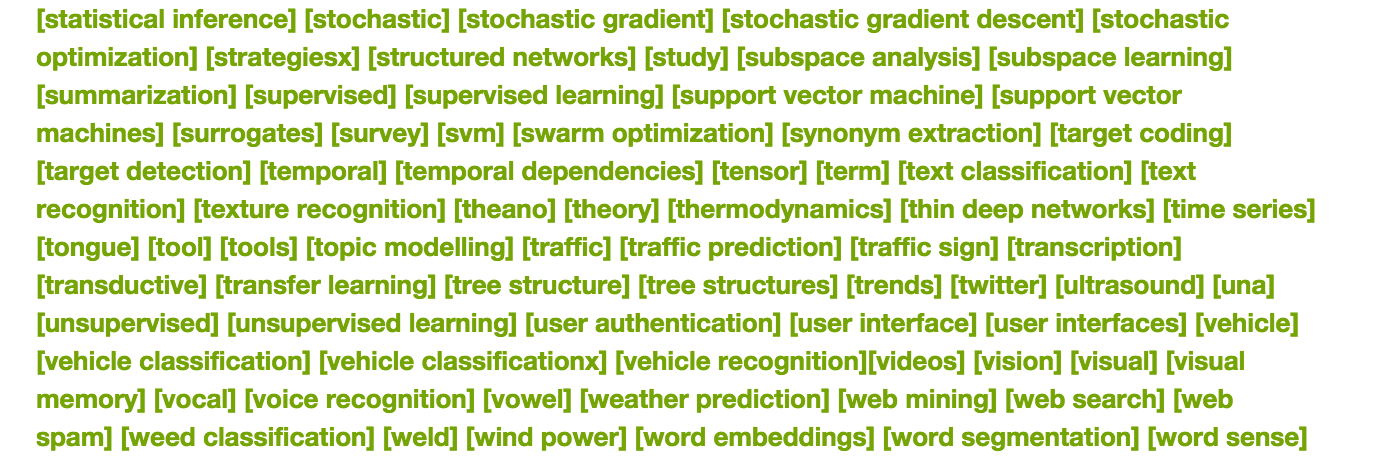}
\caption{Deeplearning.University - Keywords part 3}
\label{dluniv3}
\end{center}
\end{figure}

\begin{enumerate}
\item use FFT-based convolution - with precalculated convolution filters \cite{DBLP:journals/corr/VasilacheJMCPL14,2015convnetbenchmarks}
\item use lower resolution on floating point - in order to increase performance and support larger models (for now it uses 32 bit float or complex numbers - i.e. 2x32 bit per complex number to prepare for FFT-based convolution) \cite{DBLP:journals/corr/GuptaAGN15, 2015PeteWarden8Bits}
\item avoid copying memory between CPU and GPU more than needed \cite{2015MemkiteGPUCPUMemorySharingTveit}
  \item add support for other types of pre-trained networks than deep convolutional neural networks, e.g. recurring neural networks\cite{2015DLUnivTveit,2015DLUnivLisaLabsTveit}
\item look into more in-place calculations to save memory, i.e. supporting larger models
\item try to exploit larger parts of Metal API wrt memory layout, threadgroups to increase performance (this relates to 1.) \cite{2015AppleMetalFrameworkReference,2015AppleMetalShadingLanguageGuide,2015AppleMetalProgGuide,2015AppleMetalPerfShaders,2015AppleMetalKitReference}
\item Look into teacher-student deep networks or other compressed models for even smaller but still high quality models (recent research have shown AlexNet models being compressed from 240MB to 6.9MB), see the paper \href{http://arxiv.org/pdf/1510.00149v1.pdf}{[A Deep Neural Network Compression Pipeline]}
\item Look into algorithms for approximate matrix multiplication (i.e. convolution step speedup) to further increase speed (and reduce energy usage), interesting techniques include a) \href{http://cs.stanford.edu/people/mmahoney/cs369m/Lectures/lecture3.pdf}{[Approximating matrix multiplication and low-rank approximation]}, \href{http://arxiv.org/abs/1408.4230}{[Fast Approximate Matrix Multiplication by Solving Linear Systems]} and \href{http://down.cenet.org.cn/upfile/54/2005920141140161.pdf}{[Fast Monte-Carlo Algorithms for Approximate Matrix Multiplications]}.
\item Look into a broad set of Deep Learning applications, e.g. categories in figures \ref{dluniv1}, \ref{dluniv2} and \ref{dluniv3} from DeepLearningKit's research bibliography at \href{http://deeplearning.university}{[http://Deeplearning.University]}. It might be application specific optimizations that can be done, e.g. in the case of natural language processing with convolutional neural networks one uses 1D convolution instead of 2D (as in image classification).
\end{enumerate}

%% file: modelappstore.tex
\section{App Store for Deep Learning Models}
Given the immense asymmetry in time taken to train a Deep Learning Model versus time needed to use it (e.g. to do image recognition), it makes perfect sense to build a large repository of pre-trained models that can be (re)used several times. Since there are several popular tools used to train Deep Learning models (e.g. Caffe, Torch, Theano, DeepLearning4J, PyLearn and Nervana) we’re working on supporting importing pre-trained models in those tools into an “app store” for deep learning models (currently we’ve been primarily been working with Caffe CNN models).

\begin{figure}[h!]
\begin{center}
\includegraphics[width=12cm]{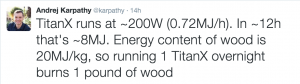}
\caption{Tweet about training model}
\label{energytweet}
\end{center}
\end{figure}

\begin{figure}[h!]
\begin{center}
\includegraphics[width=12cm]{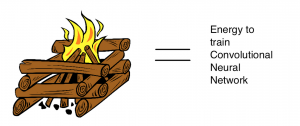}
\caption{Energy needed to train CNN}
\label{fire}
\end{center}
\end{figure}

\begin{figure}[h!]
\begin{center}
\includegraphics[width=12cm]{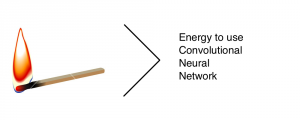}
\caption{Energy needed to run CNN}
\label{match}
\end{center}
\end{figure}

The tweet in Figure \ref{energytweet} illustrates how much energy is required to train a Deep Network (per night), some Deep Learning Models can take weeks of training on GPUs like the Nvidia TitanX, or in other words piles of wood of energy. Using a model is quite different since it requires less energy than lighting match. See figures \ref{fire} and \ref{match} for an illustration of this.

Deep Learning Models also typically have a (low) limit in the number of classes they can predict per model (e.g. in the ImageNet competition there are 1000 classes, CIFAR-100 100 classes and CIFAR-10 10 classes). This means that in order to create real-life applications one need to intelligently (and very rapid load them from SSD into GPU accessible RAM) switch between several Deep Learning Models, or if there is enough capacity one can run several models in parallel on the same GPU. Selecting an approriate Deep Learning model (i.e. which is the most likely to work well in a given context) is to our knowledge not a well-studied field of research, and in some ways it resembles the meta or universal search problem found in web search (e.g. cross-model ranking), but latency plays an even bigger part in the mobile on-device case (don’t have time to run many models). We have some ideas for a meta model for selecting a model to use, which can use input like location, time of day, and camera history to predict which models might be most relevant.

With state-of-the-art compression techniques for Convolutional Neural Network the (groundbreaking) AlexNet model from 2012 can be compressed from 240MB to 6.9MB.  This means that one could theoretically fit more than eighteen thousand AlexNet models on a 128 GB mobile device like the iPhone 6!

%% file: modelimporter.tex
\section{Deep Learning Model Importer}

Importing Deep Learning models into the model app store requires
supporting the main Deep Learning tools. The most used ones in
research are Torch and Caffe, and DeepLearningKit currently supports
converting trained Caffe models to JSON (i.e. ready to be uploaded to
app store) and then importing into Swift/Metal (or OpenCL/Vulkan with
porting) for the mobile app. Making support for importing
convolutional neural network from other tools might require getting
intimate insight into the tools, but since convolutional neural
networks are quite similar of nature the complexity and effort for
creating importers is not horrific. Proposing and supporting standards - e.g. for\
\begin{enumerate}
\item deep learning network description (i.e input to training stage)
\item input data formats (images, text, etc input to training stage)
\item trained networks (i.e. input to DeepLearningKit deep learning)
\end{enumerate}
might be a longer term goal since this will make it easier to use
pretrained models with OpenCL/Vulkan no matter which tool they are
created in.